\documentclass[10pt,twocolumn,letterpaper]{article}

\usepackage{wacv}              % To produce the CAMERA-READY version
\usepackage{graphicx}
\usepackage{amsmath}
\usepackage{amssymb}
\usepackage{booktabs}
\usepackage{multirow}
\usepackage{utfsym}
\usepackage{xcolor}
\usepackage{colortbl}

\usepackage[pagebackref,breaklinks,colorlinks]{hyperref}

\usepackage[capitalize]{cleveref}
\crefname{section}{Sec.}{Secs.}
\Crefname{section}{Section}{Sections}
\Crefname{table}{Table}{Tables}
\crefname{table}{Tab.}{Tabs.}

\def\wacvPaperID{1756} % *** Enter the WACV Paper ID here
\def\confName{WACV}
\def\confYear{2024}

\begin{document}

%%%%%%%%% TITLE - PLEASE UPDATE
\title{RPCANet: Deep Unfolding RPCA Based Infrared Small Target Detection}

\author{Fengyi Wu$^{1,}$\footnotemark[1]~~~~~~~ Tianfang Zhang$^{1,}$\footnotemark[1]~~~~~~~ Lei Li$^{2}$~~~~~~~ Yian Huang$^{1}$~~~~~~~ Zhenming Peng$^{1,}$\footnotemark[2]\\
$^{1}$University of Electronic Science and Technology of China, Chengdu, China\\
$^{2}$University of Copenhagen, Denmark\\
{\tt\small fengyiwu@std.uestc.edu.cn;}
{\tt\small sparkcarleton@gmail.com;}
{\tt\small lilei@di.ku.dk}\\
{\tt\small huangyian@std.uestc.edu.cn;}
{\tt\small zmpeng@uestc.edu.cn}
% For a paper whose authors are all at the same institution,
% omit the following lines up until the closing ``}''.
% Additional authors and addresses can be added with ``\and'',
% just like the second author.
% To save space, use either the email address or home page, not both
}
\maketitle

\renewcommand{\thefootnote}{\fnsymbol{footnote}} %将脚注符号设置为fnsymbol类型，即特殊符号表示
\footnotetext[1]{These authors contributed equally to this work.} 
\footnotetext[2]{Corresponding author.}

%%%%%%%%% ABSTRACT
\begin{abstract}
Deep learning (DL) networks have achieved remarkable performance in infrared small target detection (ISTD). However, these structures exhibit a deficiency in interpretability and are widely regarded as black boxes, as they disregard domain knowledge in ISTD. To alleviate this issue, this work proposes an interpretable deep network for detecting infrared dim targets, dubbed RPCANet. Specifically, our approach formulates the ISTD task as sparse target extraction, low-rank background estimation, and image reconstruction in a relaxed Robust Principle Component Analysis (RPCA) model. By unfolding the iterative optimization updating steps into a deep-learning framework, time-consuming and complex matrix calculations are replaced by theory-guided neural networks. RPCANet detects targets with clear interpretability and preserves the intrinsic image feature, instead of directly transforming the detection task into a matrix decomposition problem. Extensive experiments substantiate the effectiveness of our deep unfolding framework and demonstrate its trustworthy results, surpassing baseline methods in both qualitative and quantitative evaluations. Our source code is available at \url{https://github.com/fengyiwu98/RPCANet}.
% In the field of infrared small target detection (ISTD), deep learning (DL) networks have achieved remarkable performance. However, these existing structures often lack interpretability and are regarded as black boxes, as they are constructed without incorporating domain-specific traditional knowledge within ISTD. To address this limitation, we propose a model-guided deep neural network called Robust Principle Component Analysis Network (RPCANet), comprising three individual modules that emulate the processes of low-rank and sparse decomposition. Firstly, we introduce a background estimation module (BEM) that mimics the update of low-rank elements using a proximal network, eliminating the need for time-consuming matrix decompositions. Secondly, we incorporate a target extraction module (TEM) that utilizes neural layers to approximate sparsity constraint functions, replacing the conventional approach of applying soft thresholding. Finally, we merge the two updates using an image reconstruction module (IRM), which enhances the recovery process and provides guidance for target segmentation. Through end-to-end training, RPCANet effectively extracts sparse infrared targets and simulates low-rank backgrounds, resulting in improved and reliable detection performance. Extensive experiments validate the effectiveness of the framework and demonstrate the interpretability of all its modules, both in quality and quantity, when compared against baselines.
\end{abstract}

\begin{figure}
\setlength{\abovecaptionskip}{0cm}
\setlength{\belowcaptionskip}{-0cm}
\centering
	\includegraphics[scale=0.52]{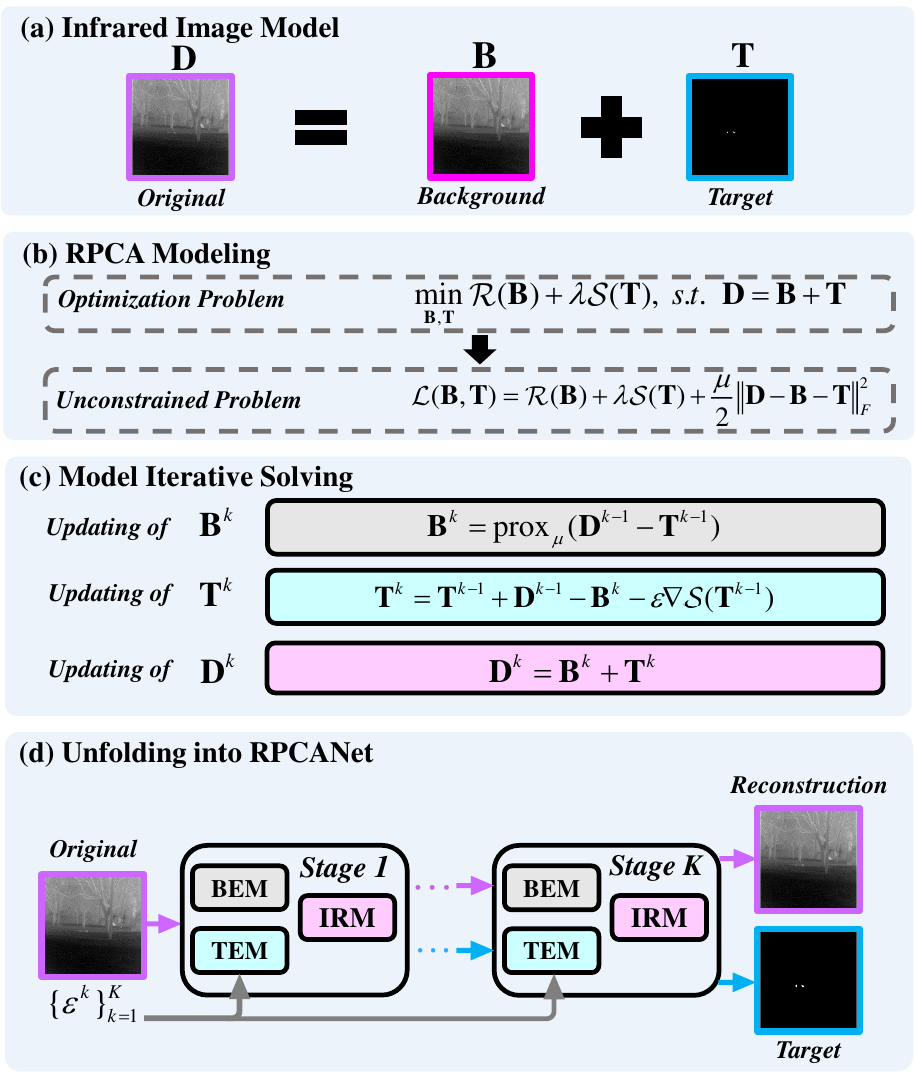}
   \caption{Overview of the suggested RPCANet. The optimization steps of the ISTD model are unfolded into a deep framework.}
\label{fig:short}
\vspace{-0.5cm}
\end{figure}
%%%%%%%%% BODY TEXT
\section{Introduction}
Research on infrared small target detection (ISTD) plays a crucial role in both civilian and military applications, such as maritime rescue \cite{prasad2017video}, early-warning systems\cite{dai2017reweighted}, and reconnaissance activities\cite{zhao2022single}. However, targets usually appear as point targets in just a few pixels due to the inherent extensive imaging distance in infrared detection, leading to limited shape and texture information. Further, background clutter and imaging system noise interfere with distinguishing infrared targets from complex noise, making ISTD a challenging academic topic.

In recent years, ISTD techniques have been advanced, falling into model-driven and data-driven categories.  The model-driven framework includes filter-based \cite{bai2010analysis, deshpande1999max, hadhoud1988two}, human vision system (HVS)-based \cite{kim2009small, chen2013local, wei2016multiscale}, and optimization-based methods \cite{gao2013infrared, wang2017infrared, dai2017reweighted}. These methods provide theoretical and physical foundations and interpretable detection results, but their efficacy often hinges on finely tuned parameters, which may not suffice when the environment changes. 

% The evolution of Artificial Neural Networks (ANNs) has enabled data-driven methods for detecting small infrared targets. While some research applies general deep-learning-based object detection techniques \cite{girshick2015fast, redmon2016you, liu2016ssd} to ISTD, the performance has been found wanting due to the significant difference between visible light and infrared in terms of shape and texture information. Thus, recent studies have treated ISTD as a segmentation problem \cite{dai2021asymmetric, dai2021attentional, li2022dense, zhang2023attention}, better capturing discriminative and salient features of small targets. However, common segmentation ISTD methods \cite{dai2021asymmetric, dai2021attentional} adopt backbones on ImageNet \cite{russakovsky2015imagenet} may constrain the infrared data representation \cite{wu2023uiu}. %Moreover, multiple downsampling [] operations in the mainstream networks lead to image feature loss, which damages both background and target information. This is undesirable given the limited size of small targets. 
%  Also, due to the small target size, the multiple downsampling operations in mainstream networks can result in target loss. As these networks are designed without ISTD domain knowledge, the reliability of their results remains questionable.

Data-driven methods for detecting small infrared targets have been enabled by the evolution of Artificial Neural Networks (ANNs). However, general deep-learning-based object detection techniques \cite{girshick2015fast, redmon2016you, liu2016ssd} have shown limited performance on ISTD due to the significant difference in shape and texture information between visible light and infrared. Therefore, recent studies have formulated ISTD as a segmentation problem \cite{dai2021asymmetric, dai2021attentional, li2022dense, zhang2023attention}, which can better capture the discriminative and salient features of small targets. However, common segmentation ISTD methods \cite{dai2021asymmetric, dai2021attentional} that adopt ImageNet \cite{russakovsky2015imagenet} backbones may limit the infrared data representation \cite{wu2023uiu}. Moreover, the multiple downsampling operations in mainstream networks can cause target loss due to the small target size. The reliability of these networks is questionable as they are designed without ISTD domain knowledge.
%The unclear interpretability in networks does not ensure that they distinguish real targets from false alarms or backgrounds. 

Taking into account the benefits and drawbacks of both model and data-driven ISTD techniques, a pertinent question arises:  \textit{Can we devise a balanced ISTD method that integrates the strengths of data-driven and model-driven methods to yield more reliable ISTD results?} Deep unfolding network (DUN), also known as algorithm unrolling, is an emerging technique that bridges the gap between iterative algorithms and neural networks. This technique has attracted a great deal of attention across multiple fields \cite{zhang2018ista, wang2020model, ren2021adaptive, tomas2023deep}. DUN constructs a network by unrolling the iterative solving algorithm of an existing model at the iteration level. Its hyperparameters are subsequently updated in a network-based fashion. By establishing systematic and precise connections between iterative algorithms and neural networks, DUNs exhibit considerable potential for building more interpretable networks\cite{an2023adversarial}. 

Interestingly, optimization-based ISTD schemes, as an extended application of image completion, involve optimization steps such as model formulation (as shown in Fig. \ref{fig:short} (a), low-rank background and sparse target) and iterative solving methods \cite{gao2013infrared,wu2023infrared}, which align with the requirement of DUNs. However, complex matrix operations hinder scholars from delving into DUN-based ISTD. Some studies in medical image restoration \cite{huang2021deep} and video separation \cite{cai2021learned} have modeled images using robust principle component analysis (RPCA). But directly learning parameters in singular value thresholding (SVT) and soft thresholding (ST) overlooks the inherent correlation of images and complicates the empirical selection of regularization parameters. %require perfect restoration of the background, relaxing the modeling of low-rank components can expedite matrix computation and alleviate the restrictions on parameter settings. 

Thus, to design an ISTD network that addresses the above issues and balances efficiency and interpretability, we develop a deep architecture called Robust Principle Component Analysis Network (RPCANet), as shown in Fig. \ref{fig:short} (d). %This optimization framework avoids complex operations such as SVT and ST. 
Rather than applying ST on deep features, this framework uses neural layers to approximate the sparsity constraint function within a target extraction module (TEM). In addition, we design a background extraction module (BEM) to mimic the proximal function with convolution layers, thus approximating the background and eliminating steps such as SVT. An image reconstruction module (IRM) is also introduced to iteratively combine the target and background outputs. %Furthermore, considering that the image residuals are more detailed, we introduce an enhanced version of RPCANet that improves reconstruction performance, named RPCANet+^+. 
In summary, our work contributes to the field of ISTD by:
\begin{itemize}
%\item We construct RPCANet as a learnable deep network architecture derived from a relaxed RPCA model, where neural network layers emulate matrix operations. RPCANet benefits from the high accuracy of data-driven networks and the superior interpretability of model-driven ISTD approaches.
\item RPCANet is a learnable deep network architecture derived from a relaxed RPCA model in Fig. \ref{fig:short} (b) and (c), combining the accuracy of data-driven networks with the interpretability of model-driven ISTD approaches.
\item Our TEM and BEM approximate the target and background for segmentation. Nonlinear proximal mapping problems for targets are handled effectively, while complex matrix calculations in background estimation are replaced with neural layers. Moreover, the proposed IRM combines the background and targets, aiming to reconstruct the image.
\item Experimental results on multiple datasets demonstrate the effectiveness of RPCANet compared to state-of-the-art baselines. Visualization of learned characteristics provides comprehensive and reliable results.
\end{itemize}

%-------------------------------------------------------------------------
\section{Related Work}

\subsection{Optimization-based ISTD}
Numerous ISTD methods have been developed over the past decades. Optimization-based methods are dominant in the model-driven category, compared to traditional filter-based \cite{bai2010analysis, deshpande1999max, hadhoud1988two} and HVS-based \cite{kim2009small, chen2013local, wei2016multiscale} schemes. The infrared patch image (IPI) model by Gao \etal \cite{gao2013infrared} introduces the method of low-rank and sparse decomposition within the stable RPCA model \cite{zhou2010stable}. This matrix-based category has been enriched by works in single \cite{zhang2018infrared, zhang2021infrared} and multi \cite{wang2017infrared} subspaces. Based on the hypothesis that tensors leverage correlation information better than matrices \cite{goldfarb2014robust}, Dai \etal proposed a reweighted infrared patch tensor (RIPT) model for ISTD \cite{dai2017reweighted}. Various strategies \cite{zhang2019infrared, kong2021infrared, wu2023infrared} based on tensors have been applied to both single and multi-frame infrared sequences. The optimization-based methods model targets and backgrounds in a mathematically interpretable manner within RPCA. However, they have limitations in robustness, parameter tuning, and efficiency\cite{zhang2022isnet, zhang2022lr}. Therefore, our goal is to develop a model that preserves interpretability while improving robustness and efficiency.

\subsection{Deep Learning-based ISTD}
In contrast to model-driven ones, convolutional neural networks (CNNs) are data-driven to learn the non-linear mappings between the original images and masks and are flexible to complicated scenarios. Wang \etal proposed a conditional GAN (MDvsFA-cGAN) \cite{wang2019miss} to reduce false alarms and missed detections in ISTD. Dai \etal presented an asymmetric contextual modulation (ACM) method \cite{dai2021asymmetric} to enhance feature fusion. Zhang \etal \cite{zhang2023attention} extracted contextual information of the target in the deep layer, based on ACM. Li \etal \cite{li2022dense} designed the DNANet with an enhanced receptive field to prevent small targets from disappearing. Wu \etal \cite{wu2023uiu} integrated residual u-blocks into the network, which preserves feature resolution. Zhang \etal \cite{zhang2022isnet} merged an edge block in a U-Net framework to enhance edge features, considering target shapes vary. Ying \etal \cite{ying2023mapping} updated labels with single-point supervision, based on effective models \cite{dai2021asymmetric, dai2021attentional, li2022dense}. However, most data-driven methods tailor the segmentation network, which is a black box, instead of incorporating classic ISTD algorithms with field knowledge. Only a few studies have tried this approach \cite{dai2021attentional, hou2021ristdnet}. Therefore, we aim to find a solution that balances interpretability and effectiveness.
\subsection{Deep Unfolding Networks} 
Deep unfolding networks (DUNs) solve iterative optimization problems via neural networks and have been widely used in image processing. Gregor and LeCun \cite{gregor2010learning} first proposed the Learned ISTA (LISTA) model, which learns the parameters of the iterative shrinkage thresholding algorithm (ISTA). Zhang \etal \cite{zhang2018ista, you2021ista} improved it by introducing ISTA-Nets, which incorporate neural layers into the update steps. Yang \etal \cite{yang2018admm} developed ADMM-CSNet, which is based on unfolding the alternating direction multiplier method (ADMM) for MRI-oriented image reconstruction. Borgerding \etal \cite{borgerding2017amp} applied the approximate message passing algorithm (AMP) to learn sparse linear inverse problems, and Zhang \etal \cite{zhang2020amp} extended it to compressive sensing. Other works also use DUNs for optimization models, such as low-rank representation \cite{zhang2022lr} and RPCA \cite{solomon2019deep, cai2021learned}. These DUNs frameworks blend the strengths of model and data-driven methods, achieving high and robust performance. Given the solid knowledge of optimization and deep learning-based methods, a progression towards DUN-based ISTD is a logical next step.
% Deep unfolding networks solve iterative optimization problems via neural networks and have been widely used in image processing. Gregor and LeCun \cite{gregor2010learning} first proposed the Learned ISTA (LISTA) model, which learns the parameters of the iterative shrinkage thresholding algorithm (ISTA). Zhang \etal improved this model by introducing ISTA-Nets \cite{zhang2018ista, you2021ista}, which incorporate neural layers into the update steps. Yang \etal \cite{yang2018admm} developed ADMM-CSNet, which is based on unfolding the alternating direction multiplier method (ADMM) for magnetic resonance imaging (MRI)-oriented image reconstruction. Mark \etal \cite{borgerding2017amp} applied the approximate message passing algorithm (AMP) to learn sparse linear inverse problems, and Zhang \etal \cite{zhang2020amp} extended this approach to compressive sensing. Several other works also solve optimization models using DUNs, such as low-rank representation \cite{zhang2022lr} and RPCA \cite{solomon2019deep, cai2021learned}. These DUNs frameworks effectively blend the strengths of model and data-driven methods, achieving a high and robust performance. Therefore, given the solid knowledge of optimization and deep learning-based methods, a progression towards DUN-based ISTD is a logical next step.

\section{Deep Unfolding RPCA Network}
\subsection{Problem Formulation}
\label{section:3.1}
For an infrared image $\mathbf{D}$, the physical model separates it into low-rank background $\mathbf{B}$ and sparse target $\mathbf{T}$ as:
\begin{equation}
\setlength{\abovedisplayskip}{4pt}
    \mathbf{D} = \mathbf{B} + \mathbf{T} \enspace,
    \vspace{-0.2cm}
\label{IPI}
\end{equation}
where $\mathbf{D}, \mathbf{B}, \mathbf{T}\in \mathbb{R}{^{m \times n}}$. In a general RPCA \cite{zhou2010stable} manner, we aim to recover the low-rank background $\mathbf{B}$ that could pair the given $\mathbf{D}$ by constraining sparse target $\mathbf{T}$ and usually transform the detection problem into:%\tf{Tell the original optimization problem, and how we approximated before. (Optimal convex functions)}
\begin{equation}
\setlength{\abovedisplayskip}{4pt}
    \min \limits_{\mathbf{B},\mathbf{T}} rank(\mathbf{B}) + \lambda \left\| \mathbf{T} \right\|_0 \quad s.t.~\mathbf{D} = \mathbf{B} + \mathbf{T} \enspace,
        \vspace{-0.2cm}
\label{RPCA}
\end{equation}
where $\lambda$ indicates a positive trade-off parameter, and ${\left\|  \cdot  \right\|_0}$ demotes the $l_0$-norm as the number of nonzero entries. 

However, solving (\ref{RPCA}) is NP-hard since the rank function and $l_0$-norm are both non-convex and discontinuous. Considering this, models like IPI \cite{gao2013infrared} individually replace them with the nuclear norm (${\left\| \cdot \right\|_*}$, the sum of singular values in a matrix) and $l_1$-norm (${\left\| \cdot \right\|_1}$) through principal component pursuit (PCP) and reformulate (\ref{RPCA}) as:
\begin{equation}
\setlength{\abovedisplayskip}{4pt}
    \mathop {\min }\limits_{\mathbf{B},\mathbf{T}} {\left\| \mathbf{B} \right\|_*} + \lambda {\left\| \mathbf{T} \right\|_1}\quad s.t.\;\mathbf{D} = \mathbf{B} + \mathbf{T} \enspace.
\label{PCP}
    \vspace{-0.15cm}
\end{equation}

In complex infrared scenarios, the background and target may vary in complexity and sparsity, and a single norm or rank function may not capture the practical constraints \cite{zhang2021infrared}. Thus, we use $\mathcal{R}(\mathbf{B})$ and $\mathcal{S}(\mathbf{T})$ to constrain the prior knowledge of the background and target images, respectively:
% In complex infrared scenarios, the background may vary in complexity, and a single nuclear norm or rank function may not adequately capture the practical constraints. Likewise, the target elements differ in sparsity, and simply utilizing l0l_0 or l1l_1-norm may also face the same issue \cite{zhang2021infrared}. Therefore, we consider a more generalized problem, where we use R(B)\mathcal{R}(\mathbf{B}) and S(T)\mathcal{S}(\mathbf{T}) to constrain the prior knowledge of the background image and the target image, respectively:%\tf{Convert Low-Rank item and Sparse item into more generalized ones. (Unknown functions)}
\begin{equation}
\setlength{\abovedisplayskip}{4pt}
    \min \limits_{\mathbf{B},\mathbf{T}} \mathcal{R}(\mathbf{B}) + \lambda \mathcal{S}(\mathbf{T}) \quad s.t.~\mathbf{D} = \mathbf{B} + \mathbf{T} \enspace.
\label{relaxPCP}
\vspace{-0.15cm}
\end{equation}

Moreover, to simplify the complexity of updating variables due to the augmented Lagrange multipliers \cite{dai2017reweighted}, we adopt a simpler and more intuitive $l_2$-norm to transform the constrained problem into an unconstrained one \cite{tseng2008accelerated} as:%\tf{Introduce why we choose such a simple Lagrange function, and shortcomings of ADMM. (Too complicated, more hyper-parameters...)}
\begin{equation}
\setlength{\abovedisplayskip}{4pt}
    \mathcal{L}(\mathbf{B},\mathbf{T}) = \mathcal{R}(\mathbf{B}) + \lambda \mathcal{S}(\mathbf{T}) + \frac{\mu}{2} \left\| \mathbf{D} - \mathbf{B} - \mathbf{T} \right\|^2_F \enspace,
    \vspace{-0.15cm}
\label{uncon}
\end{equation}
where $\mu$ is a penalty coefficient, and ${\left\| \cdot \right\|_F}$ indicates the Frobenius norm (F-norm). For a matrix $\mathbf{X}$, its F-norm equals $\sqrt {\sum\limits_{i = 1}^m {\sum\limits_{j = 1}^n {{{\left| {{\mathbf{X}_{ij}}} \right|}^2}}}}$. Based on (\ref{uncon}), we can optimize the background and target individually in an iterative scheme.

\begin{figure*}
\setlength{\abovecaptionskip}{0.1cm}
\setlength{\belowcaptionskip}{-0.1cm}
\centering
	\includegraphics[scale=0.55]{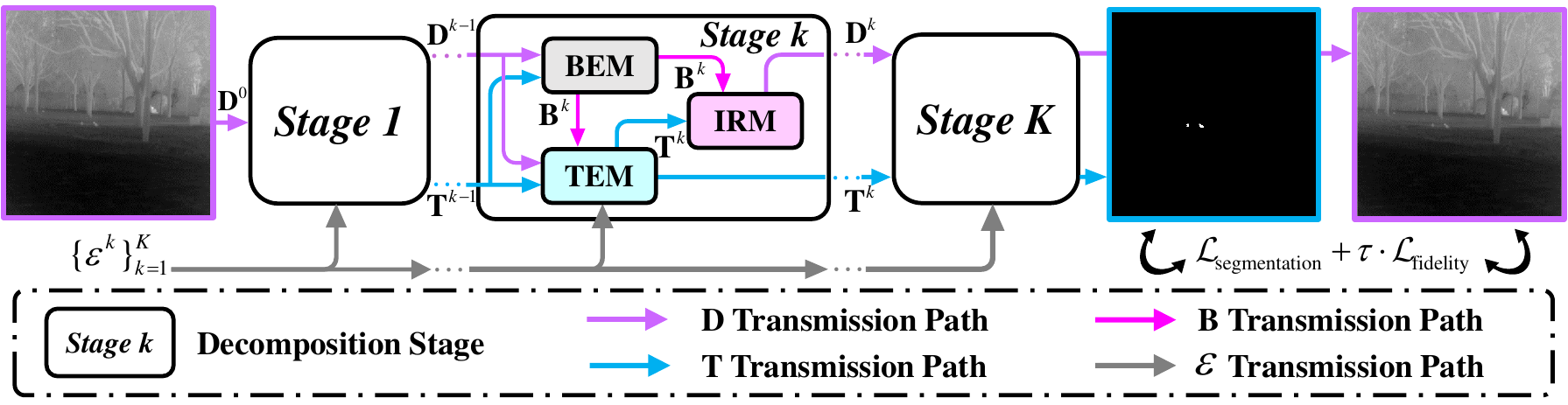}
% \begin{center}
% \fbox{\rule{0pt}{2in} \rule{.9\linewidth}{0pt}}
% \end{center}
   \caption{Overall structure of RPCANet. The network is composed of $K$ stages, each following the unfolding steps detailed in Fig. \ref{fig:detail}.}
\label{fig:overall}
\vspace{-0.3cm}
\end{figure*}

\subsection{Model Iterative Solving}
\label{section:3.2}
{\bf Updating $\mathbf{B}^*$:} To update the background, we draw the sub-problem as:
\begin{equation}
\setlength{\abovedisplayskip}{3pt}
    \mathbf{B}^* = \arg \min \limits_{\mathbf{B}} \mathcal{R}(\mathbf{B}) + \frac{\mu}{2} \left\| \mathbf{B} + \mathbf{T} - \mathbf{D}\right\|^2_F \enspace.
\label{upbg}
\vspace{-0.2cm}
\end{equation}
As (\ref{PCP}) illustrates, former optimization-based ISTD approaches usually set $\mathcal{R}(\mathbf{B})$ as ${\left\| \mathbf{B} \right\|_*}$, then degenerates (\ref{upbg}) into the sum of the nuclear norm and $l_2$ norm, and this problem has an analytical solution as:
\begin{equation}
\setlength{\abovedisplayskip}{3pt}
    \mathbf{B}^* = \mathcal{\mathcal{D}}_{\mu}(\mathbf{D}-\mathbf{T}) \enspace,
    \vspace{-0.15cm}
\label{solv}
\end{equation}
where $ \mathcal{D}_{\mu}(\cdot)$ denotes the SVT operator \cite{ma2011fixed} with the threshold of $\mu$. However, solving (\ref{solv}) involves SVD and functions on its singular values. In DUNs, we emulate this using neural networks, which means we must perform SVD on each neural tensor in each forward propagation. This poses challenges for time consumption and accuracy. \cite{cai2021learned} proposes an initialization method based on the best rank-r approximation SVD, but still faces precision issues. \cite{zhang2022lr} introduces a technique with different rank constraints for the submatrices, which improves interpretability but depends on manual tuning and ignores the intrinsic image properties when simulating matrix computation in neural layers.

Thus, instead of adopting the nuclear norm and solving it with complex SVDs, we degrade it to a constrain function $\mathcal{R}(\mathbf{B})$ in this study, and introduce a proximal operator ${\text{pro}}{{\text{x}}_\mu}( \cdot )$ to approximate the closed-form solution for the background, which is formulated as:%\tf{The upper problem is difficult to solve and tell why? (SVD, time-consuming, hard to train...)}
\begin{equation}
\setlength{\abovedisplayskip}{4pt}
    \mathbf{B}^* = {\text{prox}}_{\mu}(\mathbf{D} - \mathbf{T})\enspace.  
\label{proxB}
\end{equation}
We use convolutional layers to approximate proximal functions and solve the optimization problem. This method eliminates the need for complex matrix operations and leverages the nonlinear capabilities of neural networks to extract deep features from images in a data-driven way. And we will illustrate the detailed construction in Section \ref{section:3.3}.

{\bf Updating $\mathbf{T}^*$:} Similar with (\ref{upbg}), the sub-problem of optimizing the target is written as:
\begin{equation}
\setlength{\abovedisplayskip}{4pt}
    \mathbf{T}^* = \arg \min \limits_{\mathbf{T}} \lambda \mathcal{S}(\mathbf{T}) + \frac{\mu}{2} \left\| \mathbf{T} + \mathbf{B} - \mathbf{D} \right\|^2_F \enspace.
\label{upta}
\vspace{-0.15cm}
\end{equation}%\tf{We are facing two questions: 1) function $\mathcal{S}$ is unknown, 2) through concretization, L1 norm + L2 norm requires ISTA, shortback of ISTANet.}
As discussed in Section \ref{section:3.1}, common optimization approaches impose an $l_1$ norm constraint on the sparse target image. However, this poses the challenge of mapping the soft thresholding into the neural network \cite{zhang2018ista}. Moreover, sparse constraints often vary depending on the detection scenarios change \cite{zhang2021infrared}. Thus, we aim to derive a simpler and more intuitive representation for the closed-form solution of (\ref{upta}).
\begin{figure*}
\setlength{\abovecaptionskip}{0.1cm}
\setlength{\belowcaptionskip}{-0.1cm}
\centering
	\includegraphics[scale=0.54]{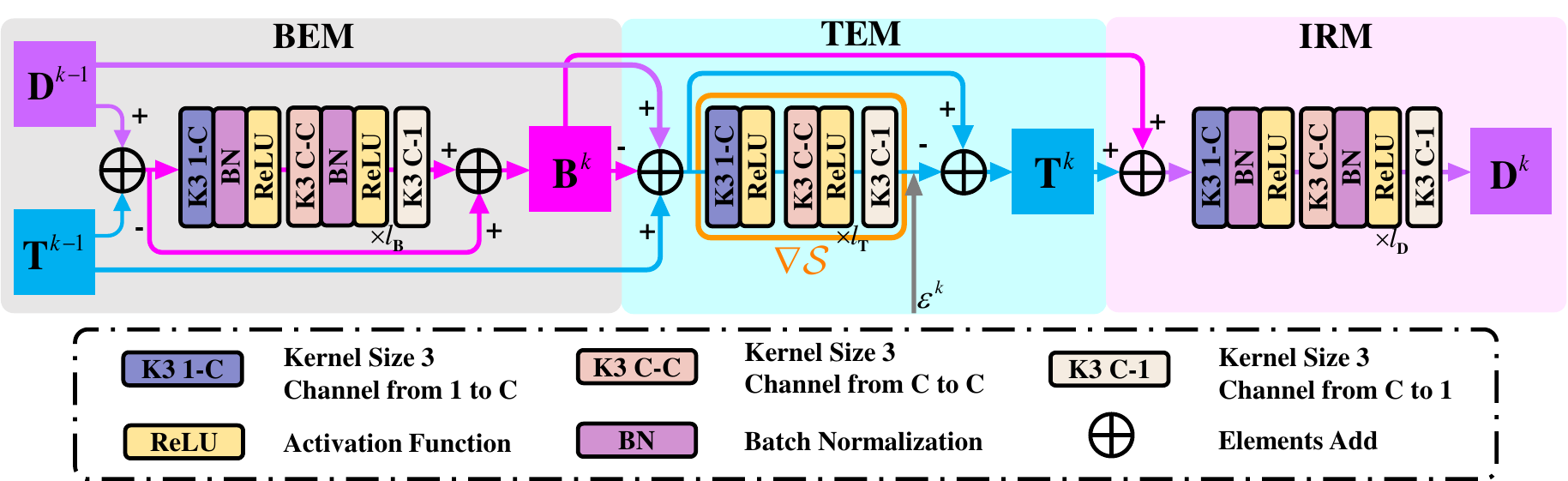}% \begin{center}
   \caption{Detail network structure of the single stage from Fig. \ref{fig:overall} in  RPCANet: background extraction module (BEM), target extraction module (TEM), and image reconstruction module (IRM).}
\label{fig:detail}
\vspace{-0.2cm}
\end{figure*}
To handle the above issues, we consider conducting Taylor expansion of $\mathcal{S}(\mathbf{T})$. For a function $f(t)$, with its Lipschitz continuous gradient function $\nabla f(t)$% (i.e., $\forall {t_1},{t_2}:\left\| {\nabla f({t_1}) - \nabla f({t_2})} \right\| \leqslant L\left\| {{t_1} - {t_2}} \right\|$, where  $L$ is a constant)
, then $f(t)$ can be Taylor approximated at the a fix point $t_0$ by:
\begin{equation}
\setlength{\abovedisplayskip}{4pt}
\hat f(t,{t_0})  \leftarrow \frac{L}{2}\left\| {t - {t_0} + \frac{1}{L}f({t_0})} \right\|^2 + C \enspace,
\vspace{-0.15cm}
\end{equation}
where $L$ is a constant, $C =  - \frac{1}{{2{L}}}{\left\| \nabla{f({t_0})} \right\|^2} + f({t_0})$ (please refer to the supplementary material for detail deductions). Based on this, we approximate $\mathcal{S}(\mathbf{T})$ at the last $\mathbf{T}^{k-1}$ as:  
\begin{equation}
\setlength{\abovedisplayskip}{4pt}
    \mathcal{\hat{S}}(\mathbf{T}, \mathbf{T}^{k-1}) \leftarrow \frac{L_{\mathcal{S}}}{2} \left\| \mathbf{T}\!-\!\mathbf{T}^{k-1}\!+\!\frac{1}{L_{\mathcal{S}}} \nabla \mathcal{S} \left( \mathbf{T}^{k-1} \right) \right\|^2_2\!+\!C_s \enspace,
\end{equation}
where $L_s$ is the Lipschitz constant of $\mathcal{S}(\mathbf{T})$ and $C_s=-\frac{1}{{2{L_s}}}{\left\| {\nabla\mathcal{S}({\mathbf{T}^{k - 1}})} \right\|_2^2} + \mathcal{S}({\mathbf{T}^{k - 1}})$ represents as a constant. And the updating formulate of the target matrix can be substituted as:
\begin{equation}
\setlength{\abovedisplayskip}{4pt}
\begin{aligned}
{\mathbf{T}^*} =& \arg \mathop {\min }\limits_\mathbf{T} \lambda \hat{\mathcal{S}}(\mathbf{T},{\mathbf{T}^{k - 1}})\!+\!\frac{\mu }{2}\left\| {\mathbf{T} + \mathbf{B} - \mathbf{D}} \right\|_F^2\\
 =& \arg \min \limits_{\mathbf{T}} \frac{{{L_\mathcal{S}}}}{2}\left\| {\mathbf{T} - {\mathbf{T}^{k - 1}} + \frac{1}{{{L_\mathcal{S}}}}\nabla \mathcal{S}({\mathbf{T}^{k - 1}})} \right\|_2^2 \\
 &+ \frac{\mu }{2}\left\| {\mathbf{T} + \mathbf{B} - \mathbf{D}} \right\|_F^2 \enspace,
\end{aligned}
\label{optT}
\end{equation}
which simply involves the sum of two $l_2$ norms, rather than a traditional $l_1$ norm constraint optimization problem, and therefore does not require the use of conventional algorithms or simulating soft thresholding \cite{zhang2018ista}. By taking the derivative of the equation and equating it to zero, a closed-form solution for updating $\mathbf{T}$ at the $k$-th step can be derived:
\begin{equation}
\setlength{\abovedisplayskip}{4pt}
\begin{aligned}
\mathbf{T}^{k} = & \frac{\lambda L_{\mathcal{S}}}{\lambda L_{\mathcal{S}} + \mu} \mathbf{T}^{k-1} + \frac{\mu}{\lambda L_{\mathcal{S}} + \mu} \left( \mathbf{D}^{k - 1} - \mathbf{B}^{k} \right)\\ 
&- \frac{\lambda}{\lambda L_{\mathcal{S}} + \mu} \nabla \mathcal{S}(\mathbf{T}^{k-1}) \enspace,
\end{aligned}
\label{closed}
\vspace{-0.15cm}
\end{equation}
where all three coefficients are constant values. By assigning each of them a learnable vector, the final equation for updating the target matrix can be reformulated:
\begin{equation}
\setlength{\abovedisplayskip}{4pt}
{\mathbf{T}^k}= \gamma {\mathbf{T}^{k-1}} + (1-\gamma )(\mathbf{D}^{k-1}-\mathbf{B}^k)-\varepsilon \nabla S({\mathbf{T}^{k - 1}})\enspace,
\label{finalt}
\vspace{-0.15cm}
\end{equation}
where $\gamma  = \frac{{\lambda {L_S}}}{{\lambda {L_S} + \mu }}$, $\varepsilon  = \frac{\lambda }{{\lambda {L_S} + \mu }}$. We learn the function $\nabla\mathcal{S}$ end-to-end without complex matrix operations such as soft thresholding, satisfying the Lipschitz continuity assumption. And the updating equation for reconstructed $\mathbf{D}^{k}$ is:
%he final update equations do not require sophisticated matrix operations such as soft thresholding but rather learn the function $\nabla\mathcal{S}$ in an end-to-end manner and meet the Lipschitz continuity assumption. And the updating equation of reconstructed image $\mathbf{D}^{k}$ can be written as:
\begin{equation}
\setlength{\abovedisplayskip}{4pt}
    {\mathbf{D}^k}={\mathbf{B}^k}+{\mathbf{T}^k}\enspace.
    \label{finald}
\vspace{-0.1cm}
\end{equation}
% \begin{equation}
% \left\{
% \begin{aligned}
% {\mathbf{B}^k}&= {\text{pro}}{{\text{x}}_\mu }({\mathbf{D}^{k - 1}} - {\mathbf{T}^{k - 1}})\\ 
% {\mathbf{T}^k}&= \gamma {\mathbf{T}^{k-1}} + (1-\gamma )(\mathbf{D}^{k-1}-\mathbf{B}^k)-\varepsilon \nabla S({\mathbf{T}^{k - 1}})\\
% {\mathbf{D}^k}&={\mathbf{B}^k}+{\mathbf{T}^k}
% \end{aligned}
% \right. .
% \label{finalup}
% \end{equation}
%\tf{In summary, there are two key update progress: B and T}
\subsection{RPCANet Framework}
\label{section:3.3}
This section describes the overall architecture of the network and module design of RPCANet, based on the optimization equations in Section \ref{section:3.2}. As shown in Fig. \ref{fig:overall}, the input to the network is an infrared image $\mathbf{X} \in \mathbb{R}^{H\times W}$ with targets, where $H$ and  $W$ are the image height and width, and the update parameters are initialized as $\mathbf{D}^0 = \textbf{X}$ and $\mathbf{T}^0 = 0$. These parameters are then passed through $K$ decomposition stages, each corresponding to an iterative matrix low-rank sparse decomposition process, to simulate the update operation of multiple iterations in model-driven approaches.

In detail, the updated parameters $\mathbf{D}^{k-1}$ and $\mathbf{T}^{k-1}$ are fed into the $k$-th decomposition stage, where $k\in\{1, \dots, K\}$. The background $\mathbf{B}^k$, target $\mathbf{T}^k$, and reconstructed result $\mathbf{D}^k$ for the current stage are estimated by BEM, TEM, and IRM, respectively. Typically, $\mathbf{B}^k$ represents the latent variable of the current decomposition stage and is not involved in the parameter transfer between stages.
%and three modules are empolyed: Background Estimation Module (BEM), Target Extraction Module (TEM) and Image Reconstruction Module (IRM). These modules estimate 

{\bf Background Estimation Module (BEM):} As shown in Fig. \ref{fig:detail}, we adopt BEM to estimate the background. In (\ref{proxB}), proximal operator ${\text{pro}}{{\text{x}}_\mu}( \cdot )$ is to-be-decided. Here, inspired by former DUN-based works \cite{wang2023indudonet+, wang2020model,tomas2023deep, zhang2020amp} and in the spirits of flexibility \cite{you2021ista}, we adopt a residual structure ${\text{proxNe}}{\text{t}}(\cdot)$ to simulate this operator. As (\ref{Bnet}) formulates:  %where $C$ is the number of increased channels while $l_\mathbf{B}$ is that of convolution layers. %Each convolutional layer employs a kernel size of $3\times3$ and a stride of 1, with the number of channel $C$ being set as 32.
    \begin{equation}
    \setlength{\abovedisplayskip}{4pt}
    \begin{aligned}
 {\mathbf{B}^k} &= {\text{ proxNet}}({\mathbf{D}^{k - 1}} - {\mathbf{T}^{k - 1}})  \\
  &= {\mathbf{D}^{k - 1}} - {\mathbf{T}^{k - 1}} + {{\mathcal{F}}^k}({\mathbf{D}^{k - 1}} - {\mathbf{T}^{k - 1}}) \enspace ,
    \end{aligned}
    \vspace{-0.15cm}
    \label{Bnet}
    \end{equation}
where ${\mathcal{F}}^k( \cdot )$ indicates the $3\times3$ convolution group in the structure as Fig. \ref{fig:detail} shows.
%which write as ${{\mathcal{F}}^k}( \cdot ) = {\mathcal{F}}_{rec}^k({\mathcal{F}}_{CB,l_{\bf{B}}}^k( \cdots {\mathcal{F}}_{CB,1}^k({\mathcal{F}}_{ext}^k( \cdot))))$.
Here, ${\mathcal{F}}^k( \cdot )$ consists of $l_{\mathbf{B}}$ middle layers $[Conv + BN +ReLU]$, and two convolution layers: feature extraction layer $[Conv + BN + ReLU ]$ and image reconstruction layer $[Conv]$, respectively. Here, BN stands for batch normalization and ReLU for rectified linear unit \cite{nair2010rectified}. All of them are in the stride of 1 and the padding of 1, and we set $C = 32$ and $l_{\mathbf{B}}=3$ in this study.
\begin{figure}[t]
\setlength{\abovecaptionskip}{0.1cm}
\setlength{\belowcaptionskip}{-0.4cm}
    \centering
    \includegraphics[scale=0.215]{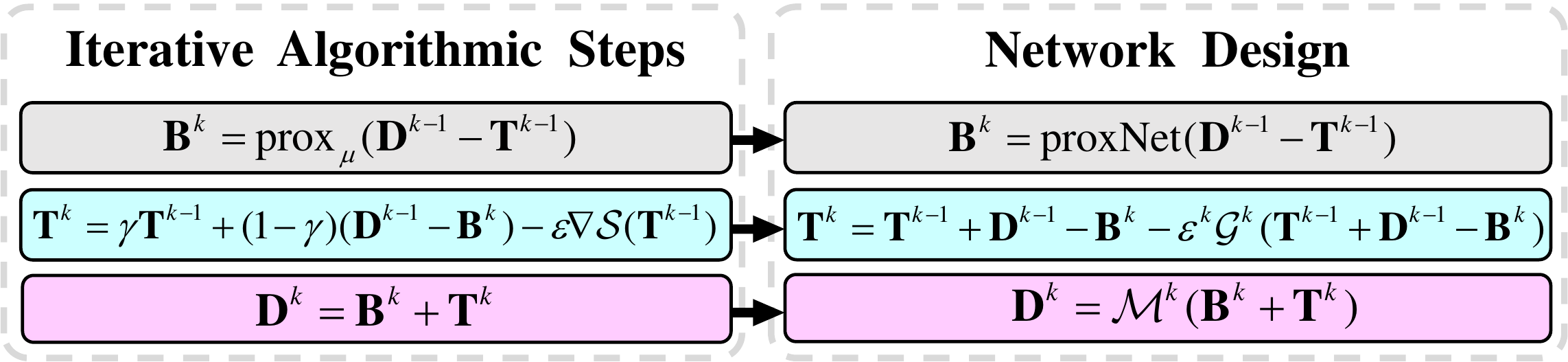}
    \caption{Relationship between the steps of the optimization algorithm and the modules of the deep unfolding network.}
    \vspace{-0.2cm}
    \label{fig:corres}
\end{figure}
\begin{figure*}[t]
\setlength{\abovecaptionskip}{0cm}
\setlength{\belowcaptionskip}{-0.1cm}
    \centering
    \includegraphics[scale=0.222]{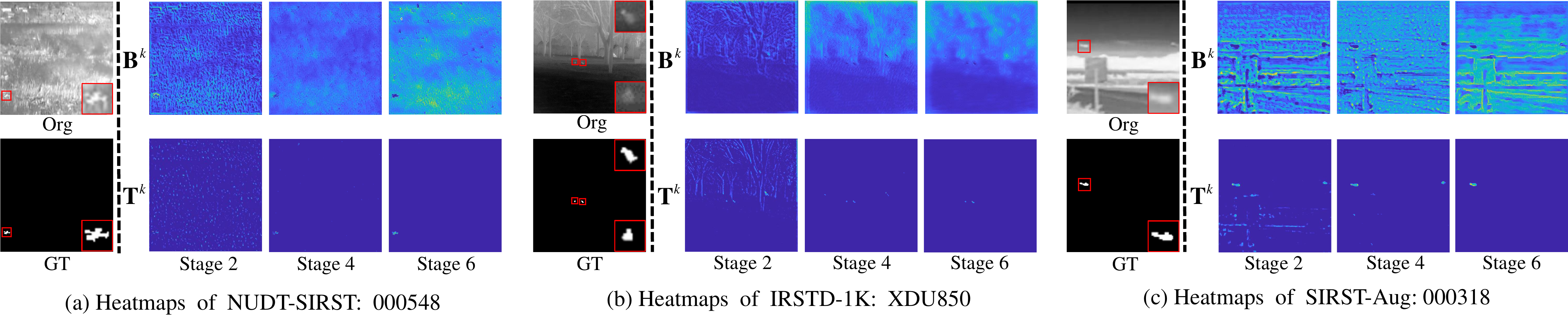}
    \caption{Heatmaps of variables $\mathbf{B}^k$ and $\mathbf{T}^k$ in different decomposition stages of RPCANet on three representative scenes when $K=6$.}
    \label{fig:unfolding}
    \vspace{-0.2cm}
\end{figure*}

{\bf Target Extraction Module (TEM):} This module takes the updated background $\mathbf{B}^k$, target $\mathbf{T}^{k-1}$, and reconstruction result $\mathbf{D}^{k-1}$ as inputs, as shown in Fig. \ref{fig:detail}. Moreover, we set $\gamma = 0.5$ to evenly treat the three inputs and  rewrite (\ref{finalt}) as:
\begin{equation}
\setlength{\abovedisplayskip}{4pt}
{\mathbf{T}^k}\!=\!{\mathbf{T}^{k - 1}}\!+\!{\mathbf{D}^{k - 1}}\!-\!{\mathbf{B}^k}\!-\!{\varepsilon}\nabla\mathcal{S}({\mathbf{T}^{k - 1}}) \enspace.
\vspace{-0.15cm}
\end{equation}
We assign the parameter learning task to $\epsilon$ as $\epsilon^k$, which is a learnable scalar independent of each reconstruction stage and does not share parameters. As to the Lipschitz continuous gradient function $\nabla\mathcal{S}$, \cite{virmaux2018lipschitz} finds that a single-layer CNN consisting of a convolution layer and a ReLU activation function is Lipschitz continuous, which also holds for multiple stacked layers. Therefore, to avoid complex network design, we adopt simple convolution layers and ReLU to simulate the function $\nabla\mathcal{S}$ as shown in the orange box in Fig. \ref{fig:detail}, in accordance with the assumption in Section \ref{section:3.2}. We also introduce the difference of last ${\bf{D}}^{k - 1}$ and updated $\mathbf{B}^k$ to enhance successive target feature, and the update function of $\mathbf{T}^k$ is written as:
\begin{equation}
\setlength{\abovedisplayskip}{4pt}
{{\bf{T}}^k\!=\!{{\bf{T}}^{k\!-\!1}}\!+\!{{\bf{D}}^{k\!-\!1}}\!-\!{{\bf{B}}^k}\!-\!{\varepsilon ^k}{{\mathcal{G}}^k}({{\bf{T}}^{k\!-\!1}}\!+\!{{\bf{D}}^{k\!-\!1}}\!-\!{{\bf{B}}^k})}.
\label{tnet}
\end{equation}
Specifically, ${\mathcal{G}}^k( \cdot )$ comprises an initial convolution layer, $l_{\mathbf{T}}$ of middle layers, and a reconstruction layer. Since the Lipschitz continuous property of the statistical operation BN is not clear \cite{virmaux2018lipschitz}, we omit the BN from the convolution block.
%simulated by the neural network is shown in the orange box in Fig. \ref{fig:detail}, with $l_T$ convolution layers. The updated target image $T^k$ is the output of TEM.
%Virmaux \etal \cite{virmaux2018lipschitz} explored the Lipschitz continuity property of neural networks and propose the Lipschitz constant estimation methods for multilayer perceptrons (MLP) and CNNs, which offers a solid basis for applying this attribute. 

{\bf Image Reconstruction Module (IRM):} To align with the RPCA process that maps the target and background into a restored image, we devise an IRM that converts the decomposition task into an image reconstruction task with a neural network  ${\mathcal{M}^k( \cdot )}$, as presented in Fig. \ref{fig:detail}:
\begin{equation}
\setlength{\abovedisplayskip}{4pt}
{{\bf{D}}^k} = {\mathcal{M}^k}({{\bf{B}}^k} + {{\bf{T}}^k})\enspace.
\vspace{-0.15cm}
\end{equation}
Instead of applying residual blocks or other complex networks in the reconstruction module \cite{wang2020model,wang2023indudonet+}, we employ a simple and decent CNN architecture \cite{zhang2018ffdnet} in learning image features and mapping the decomposed background and target effectively. Similar to ${\mathcal{F}}^k( \cdot )$, ${\mathcal{M}^k( \cdot )}$ has three types of convolution layer with $l_{\mathbf{D}}$ middle layer, where $l_{\mathbf{D}} =3$.

The corresponding updating operations in optimization-based and our DUN-based frameworks are shown in Fig. \ref{fig:corres}. To sum up,  we propose an end-to-end training framework for the ISTD task named RPCANet. Its trainable parameters set $\Theta$ includes the convolutional network parameters in BEM, TEM, and IRM in each decomposition stage, and the independent trainable scalar $\epsilon$ in TEM, which collects as $\Theta  = \left\{ {\Theta _{\mathbf{BEM}}^k,\Theta _{\mathbf{TEM}}^k,\Theta _{\mathbf{IRM}}^k,{\varepsilon ^k}} \right\}_{k = 1}^K$, where $K$ is the number of overall decomposition stages.

% Please add the following required packages to your document preamble:
% \usepackage{multirow}
\section{Network Training}
\subsection{Training Loss}
Since we separate an ISTD task into the target segmentation and infrared image reconstruction, thus the loss function consists of two components: $\mathcal{L}_\text{segmentation}$ and $\mathcal{L}_\text{fidelity}$. The former measures the target segmentation performance using SoftIoU \cite{rahman2016optimizing}, while the latter measures the infrared image reconstruction performance using the least squares error between the reconstructed image and the original image. The loss function is defined as follows:
\begin{equation}
\setlength{\abovedisplayskip}{3pt}
\begin{aligned}
&{{\mathcal{L}}_{{\text{total}}}} = {\mathcal{L}_\text{segmentation} + \tau  \cdot {\mathcal{L}_\text{fidelity}}} \\
&=1 \!-\!\frac{1}{{{N_t}}}\sum\limits_{i = 1}^{{N_t}}{\frac{{TP}}{{FP\!+\!TP\!+\!FN}}}\!+\!\frac{ \tau }{{{N_t}N}}\sum\limits_{i = 1}^{{N_t}} {\left\| {{\mathbf{D}^K}\!-\!\mathbf{D}} \right\|_F^2}, 
\end{aligned}
\end{equation}
where $N_t$ and $N$ are the total training number and total pixels per image. $\tau$ is the regularization parameter and set to 0.01 in our experiment.
\begin{figure*}[t]
\setlength{\abovecaptionskip}{0.cm}
\setlength{\belowcaptionskip}{-0.2cm}
    \centering
    \includegraphics[scale=0.40]{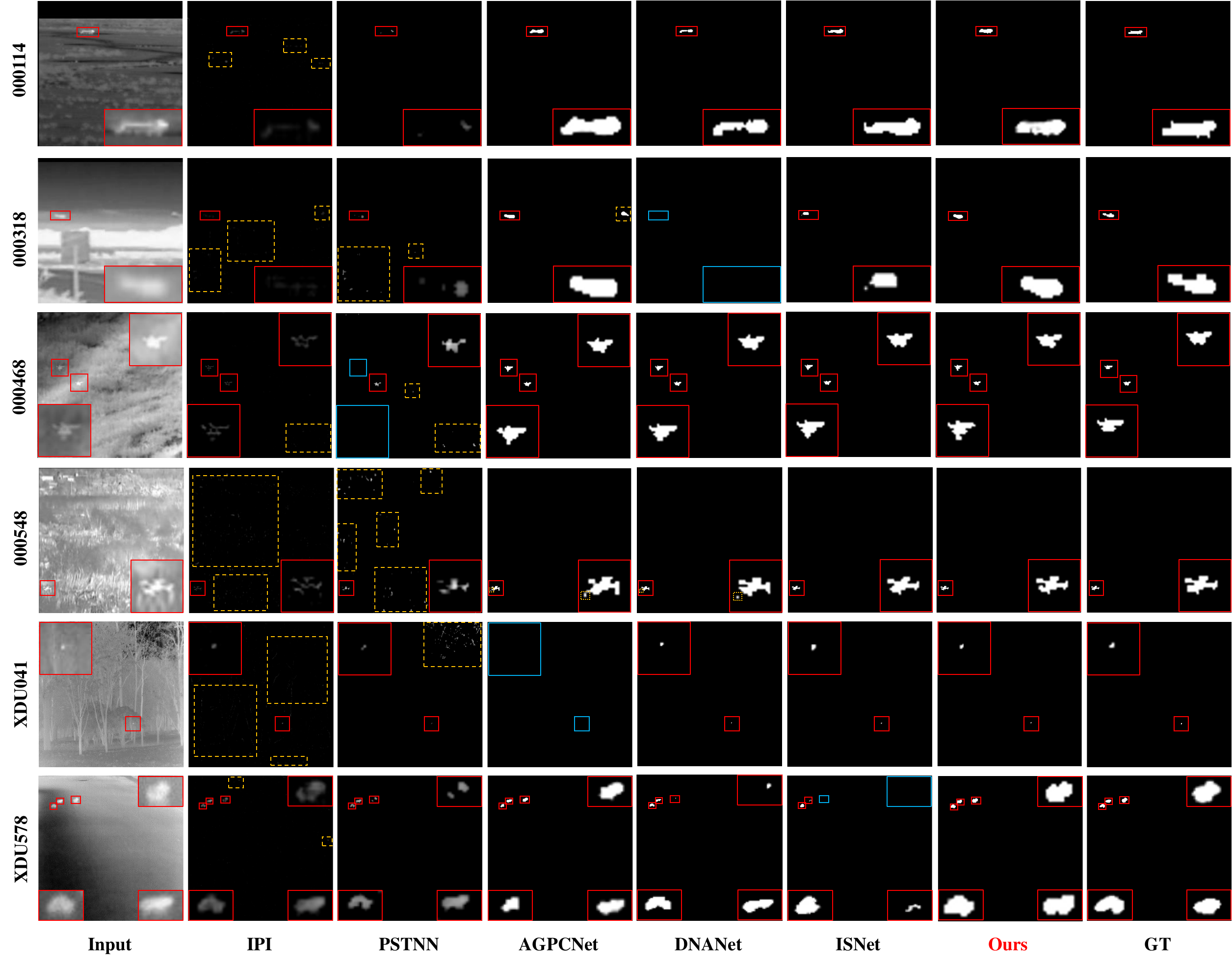}
    \caption{Representative visual results from various ISTD methods. Accurately detected targets, missed targets, and false alarms are each highlighted by boxes in red, blue, and yellow.}
    \label{fig:pres}
    \vspace{-0.25cm}
\end{figure*}
\subsection{Implementation Details}
We conduct experiments on three publicly available datasets: SIRST-Aug \cite{zhang2023attention}, NUDT-SIRST \cite{li2022dense}, and IRSTD-1K \cite{zhang2022isnet}, with their split strategies, all images are normalized into $256\times256$. Besides, evaluation metrics are two target-level: probability of detection ($P_d$) and false alarm rate ($F_a$); two pixel-level: mean intersection over union ($mIoU$) and F-measure ($F_1$); and the receiver operating characteristic (ROC) curve and the area under ROC (AUC).

We train our model for 400 epochs in the PyTorch framework on each dataset using an Nvidia GeForce 3090 GPU. We use an Adam optimizer with an initial learning rate of $10^{-4}$ in the poly policy, where the initial learning rate is multiplied by ${\left( 1-\frac{iter}{total\_iter} \right)}^{0.9}$ and a batch size of 8.
%------------------------------------------------------------------------
\section{Experiment}
To confirm the fundamental process of the suggested network, we verify the model by visualizing mid-layers and conduct ablation studies. Next, we describe experiments on open-source datasets for performance evaluation.
\begin{table}[t]
\caption{Effect of stage number $K$ on the detection performance in $mIoU$ ($\%$), $F_1$($\%$), $P_d$ ($\%$), and $F_a$ ($10^{-5}$) on SIRST-Aug \cite{zhang2023attention}.}
\small
\centering
\begin{tabular}{c|c|cccc}
\hline
Stages ($K$) &Params& $mIoU$ $\uparrow$ & $F_1$ $\uparrow$ & $P_d$ $\uparrow$ & $F_a$  $\downarrow$ \\ \hline
1    & 0.113M& 60.10 &   75.08   &  98.21  &  62.20     \\ 
2     &  0.227M&  69.26    &  81.84  &  \textbf{98.49}  &  46.12  \\ 
3      & 0.340M &   69.82   &  82.23  &  98.07  &  42.63  \\
4      & 0.453M &    69.50  & 82.01   &   98.07 &  40.36  \\ 
5      & 0.567M &    70.24  &   82.52 &  96.01  & 36.02   \\ 
\rowcolor{gray!20}6      & 0.680M  &    \textbf{72.54 }     &   \textbf{84.08 }   &  98.21     &  \textbf{34.14}   \\ 
7      & 0.793M &   70.98   &  83.02  &  96.15  &  35.85  \\ \hline
\end{tabular}
\label{paramK}
\vspace{-0.25cm}
\end{table}
\begin{table}[t]
\caption{Effect of layer number $l_\mathbf{T}$ on the detection performance in $mIoU$ ($\%$), $F_1$ ($\%$), $P_d$ ($\%$), and $F_a$ ($10^{-5}$) on SIRST-Aug \cite{zhang2023attention}.}
\centering
\small
\begin{tabular}{c|c|cccc}
\hline
TEM  &   Params& $mIoU$ $\uparrow$ & $F_1$ $\uparrow$ & $P_d$ $\uparrow$ & $F_a$ $\downarrow$  \\ \hline
 $l_\mathbf{T}=1$   &    0.402M      & 67.07& 80.29  & \textbf{99.17}   & 40.53   \\ 
 $l_\mathbf{T}=3$     &   0.513M    &  70.63  &  82.79  &  97.80&41.36  \\ 
$l_\mathbf{T}=9$    &   0.846M      &  67.99 &80.95&  94.36  & \textbf{32.75}   \\
$l_\mathbf{T}=12$    &   1.013M      &  69.98 &82.34&  97.25  & 38.46  \\ \hline
\rowcolor{gray!20}$l_\mathbf{T}=6$  &   0.680M      &   \textbf{72.54}      &   \textbf{84.08}    & 98.21     &  34.14   \\\hline
\end{tabular}
\label{paraml}
\vspace{-0.25cm}
\end{table}
%------------------------------------------------------------------------
\subsection{Model Verification}
Fig. \ref{fig:unfolding} presents the heatmaps of the intermediate update variable $\mathbf{B}^k$ and $\mathbf{T}^k$, where $k\in\{2,4,6\}$. At lower stages, the network prioritizes low-level edge texture information in the background, leading to insufficient background reconstruction. As the decomposition deepens, the network acquires more comprehensive information, demonstrating its ability to learn from formula-guided training. 

Likewise, $\mathbf{T}^k$ progressively learns the target's location from disorganized high-frequency information, resulting in a sparse matrix as the segmented output. Fig. \ref{fig:unfolding} (c) shows that false alarms in the initial stages are eliminated under mask supervision, showing the data-driven way corrects the detection result. These successive heatmaps verify the effectiveness of RPCANet in learning infrared variables according to training data while preserving the interpretability of RPCA model that facilitates the ISTD task.
\begin{table}[t]
\caption{Studies on different proximal networks and w/wo IRM on the detection performance in $mIoU$ ($\%$), $F_1$ ($\%$), $P_d$ ($\%$), and $F_a$ ($10^{-5}$) on SIRST-Aug \cite{zhang2023attention}.}
\centering
\small
\resizebox{\linewidth}{!}{
\begin{tabular}{cc|c|cccc}
\hline
BEM & IRM &  Params& $mIoU$ $\uparrow$ & $F_1$ $\uparrow$ & $P_d$ $\uparrow$ & $F_a$ $\downarrow$  \\ \hline
 RB   &  $\usym{2713}$   &  0.624M &  66.33 & 79.76   &  98.07  & 61.39   \\ 
 CNN   &  $\usym{2713}$   & 0.679M  &71.43 & 83.33&97.77 & \textbf{32.45}   \\ \hline
Ours &  $\usym{2717}$  & 0.507M&   67.07   &  80.29  &  96.56  & 36.89   \\ \hline
 \rowcolor{gray!20}Ours &  $\usym{2713}$   &  0.680M    & \textbf{72.54 }     &   \textbf{84.08}    &  \textbf{98.21}     &  34.14   \\\hline
\end{tabular}
}
\label{ablation}
\vspace{-0.3cm}
\end{table}
\setlength{\abovedisplayskip}{2pt}
\begin{table}[]
\caption{Studies on $\nabla\mathcal{S}$ simulation network in detection performance $mIoU$ ($\%$), $F_1$ ($\%$) on three different datasets.}
\small
\resizebox{\linewidth}{!}{
\begin{tabular}{c|cc|cc|cc}
\hline
\multirow{2}{*}{Config} & \multicolumn{2}{c|}{NUDT-SIRST \cite{li2022dense}} & \multicolumn{2}{c|}{IRSTD-1K \cite{zhang2022isnet}} & \multicolumn{2}{c}{SIRST-Aug \cite{zhang2023attention}} \\ \cline{2-7} 
               & mIoU$\uparrow$& F1$\uparrow$ & mIoU$\uparrow$ & F1$\uparrow$ & mIoU$\uparrow$  & F1$\uparrow$ \\ \hline
  $ {{\bf{T}}^{k - 1}} $             & 88.54 &93.92&60.98&75.76& 70.54& 82.72\\ \hline
 \rowcolor{gray!20}Ours &\textbf{89.31}&\textbf{94.35} &\textbf{63.21}&\textbf{77.45}&       \textbf{72.54}        &      \textbf{84.08}        \\ \hline
\end{tabular}
}
\vspace{-0.4cm}
\label{ablsingle}
\end{table}
\begin{table*}[t]
\setlength{\abovedisplayskip}{1pt}
\caption{$mIoU$ ($\%$), $F_1$ ($\%$), $P_d$ ($\%$), $F_a$ ($10^{-5}$), and runtime values of different methods performance on NUDT-SIRST \cite{li2022dense}, IRSTD-1K \cite{zhang2022isnet}, and SIRST-Aug \cite{zhang2023attention}. The second column records the parameter statistics for data-driven algorithms.}
\centering
\resizebox{\linewidth}{!}{
\begin{tabular}{l|c|cccc|cccc|cccc|c}
\hline
\multirow{2}{*}{Methods} & \multirow{2}{*}{Params} & \multicolumn{4}{c|}{NUDT-SIRST \cite{li2022dense}} & \multicolumn{4}{c|}{IRSTD-1K \cite{zhang2022isnet}} & \multicolumn{4}{c|}{SIRST-Aug \cite{zhang2023attention}}&Time (s) \\ \cline{3-14} 
                          &                         &$mIoU$ $\uparrow$ & $F_1$ $\uparrow$ & $P_d$ $\uparrow$ & $F_a$ $\downarrow$    & $mIoU$ $\uparrow$ & $F_1$ $\uparrow$ & $P_d$ $\uparrow$ & $F_a$ $\downarrow$     & $mIoU$ $\uparrow$ & $F_1$ $\uparrow$ & $P_d$ $\uparrow$ & $F_a$ $\downarrow$   & CPU/GPU\\ \hline
Tophat\cite{bai2010analysis}       &    -     &   22.23   &    36.37     &    96.19  &   76.27    &   5.35    &  10.15       &  68.73     & 86.50 &  16.70  &  28.62  &  95.05    &  11.77  &  0.0111/-      \\ 
MPCM \cite{wei2016multiscale} &  -     &    9.26    &   16.95    &   70.58    &  32.74     &  14.87       &  25.89    &  68.73    &  6.51 &     19.76   &   33.00    &   93.40    &  3.14  & 0.0624/-  \\ 
IPI    \cite{gao2013infrared}       &   -     &     34.62    &    51.43   &   92.38    &   7.54    &   18.64      &    31.42   &   78.01   &   11.10 &    21.93     &  35.87     &      80.06      &   1.62 & 3.0972/-   \\
%NRAM    \cite{gao2013infrared}       &   -     &     31.72    &    48.17   &   95.34    &   5.57    &   13.77      &    24.21   &   83.17   &   6.41 &    19.00     &  31.93     &      94.64      &   3.36 & 3.0972/-   \\
PSTNN   \cite{zhang2019infrared}        &    -    &  25.46       &  40.58     &   78.52   &     7.95  &    16.38     &  28.15     &   69.07   & 7.65 &   13.83      &   24.30    &    59.97   &  \textbf{1.56}  & 0.2249/-  \\ \hline
ACM  \cite{dai2021asymmetric} &  0.398M   &    69.00     &   81.66    &   95.98    &   13.34    &   61.56      &  76.20     & 92.93     &   8.88  &  70.49    &   82.62    &     96.70  &   35.29 & -/0.0072 \\ 
ALCNet \cite{dai2021attentional}    &  0.427M      &  71.48       &  83.37     &   96.30    &  11.45     &    58.23     &    73.61   &   92.92   &  10.31 &      66.21   &  79.67     &  97.80     & 37.40 &  -/0.0070  \\ 
ISNet   \cite{zhang2022isnet}       &    0.967M   &   87.51      &  93.34     &   97.35    &  3.37     &  55.29       &    71.21   &  \textbf{94.61}    &  14.19  &     70.51   &  82.71     &   97.66    &  31.57   & -/0.0132 \\ 
%RDIANet    \cite{TGRS23RDIAN}     &  0.217     &  70.34       &   82.59    &   99.31    &   38.48   &     84.69   &   91.70    &    97.78   &  5.34     &    62.27     &   76.52    & 91.58     &   6.59   \\ 
AGPCNet    \cite{zhang2023attention} &  12.360M    &    85.40     &   92.13    &   98.10    &   4.72    &      61.00   &    75.76   &   89.35   &  5.34   & 72.16     &  83.83     &    \textbf{99.03}   &  35.56  &  -/0.0205  \\ 
DNANet   \cite{li2022dense}              &  4.697M   &   83.94   &    91.27     &    \textbf{98.52}   &    6.21   &   60.51    &     75.40    &   91.07    &  5.43    &     69.58   &    82.06     &     96.14  &   27.26   &-/0.0250      \\ 
UIUNet    \cite{wu2023uiu}    &   50.540M    &    88.71  &   94.01    &     91.43  &    \textbf{1.89 }  &     63.06    &    77.35   &  93.60    &  6.57   &    71.80     &   83.59    &    98.35   &  28.29  & -/0.0261  \\ \hline
	\rowcolor{gray!20} Ours   &   0.680M     &    \textbf{89.31}    &  \textbf{94.35}     &   97.14    &    2.87   &     \textbf{63.21}   &   \textbf{77.45}    &  88.31    & \textbf{4.39}  &   \textbf{72.54}      &   \textbf{84.08}    &  98.21     &  34.14 & -/0.0096  \\ \hline
\end{tabular}
}
\label{baseline}
\vspace{-0.3cm}
\end{table*}

\subsection{Ablation Studies}
{\bf Effects of Parameter $K$ and $l_\mathbf{T}$:}
Table \ref{paramK} compares the effects of different reconstruction stage counts. Our method only reaches significant detection performance in two stages, validating the essential attributes of the suggested RPCANet. Additionally, we see that $K=7$ has a worse ISTD performance than $K = 6$, which makes sense given that a higher K could hinder gradient propagation, based on this, we set $K$ to 6. Similarly, as shown in Table \ref{paraml}, the number of the convolutional layer $l_\mathbf{T}$ faces the same manner, the performance of the network can be improved as $l_\mathbf{T}$ raises, but the gain effect of too many reconstruction stages on the network performance is limited, thus, we take $l_\mathbf{T}=6$ in all our experiments.

{\bf Studies of $\text{proxNet}(\cdot)$ and IRM:} The emulation of proximal operators acts an essential role in background estimation\cite{wang2020model,you2021ista}, we investigate two $\text{proxNet}(\cdot)$ constructions (residual block (RB) \cite{wang2023indudonet+} and plain CNN) for proximal operators. Table \ref{ablation} demonstrates that RB can yield some effects and a small parameter count, our network in Section \ref{section:3.3} achieves better results with a minimal parameter increase. We also examine the influence of the IRM. As depicted in the third row of Table \ref{ablation}, RPCANet, guided by the reconstruction module, exhibits substantial improvements in four metrics, proving the module's effectiveness. 

{\bf Studies of $\nabla\mathcal{S}$ Simulation Network:} We investigate the impact of incorporating ${\bf{D}}^{k - 1}$ and $\mathbf{B}^k$ subtraction within ${\mathcal{G}}^k( \cdot )$ for target feature enhancement. Table \ref{ablsingle} compares two configurations. While using ${\bf{T}}^{k - 1}$ alone yields satisfactory results, incorporating additional information in (\ref{tnet}) significantly improves $mIoU$ and $F_1$. We can conclude that appropriate modifications to the network structure, guided by the model's prior, can enhance detection performance.

\subsection{Comparison to State-of-the-art Methods}
For model-driven methods, we select: filter-based Tophat \cite{bai2010analysis}, HVS-based MPCM \cite{wei2016multiscale}, matrix optimization-based IPI \cite{gao2013infrared}, and tensor optimization-based PSTNN \cite{zhang2019infrared}. For data-driven algorithms, we conduct experiments on ACM \cite{dai2021asymmetric}, ALCNet \cite{dai2021attentional}, ISNet \cite{zhang2022isnet}, AGPCNet \cite{zhang2023attention}, DNANet \cite{li2022dense}, and UIUNet \cite{wu2023uiu}.

{\bf Qualitative Results:} Fig. \ref{fig:pres} depicts visual results obtained from various algorithms applied to three datasets. RPCANet effectively generalizes complex scenarios, producing outputs with accurate target shapes and low false alarm rates. Compared to model-driven algorithms, our network excels at suppressing false alarms while preserving the sparse shape of infrared targets. Although some DL-based models can effectively reduce false alarms, they may struggle in small and complex situations, leading to missed detections. Algorithms like ISNet largely maintain the target shape; however, they occasionally encounter difficulties in multi-target scenarios. The visualization results clearly illustrate that our network combines the advantages of optimization-based and DL-based methods.

\begin{figure}[t]
\setlength{\abovecaptionskip}{0.1cm}
\setlength{\belowcaptionskip}{-0.3cm}
    \centering
    \includegraphics[scale=0.55]{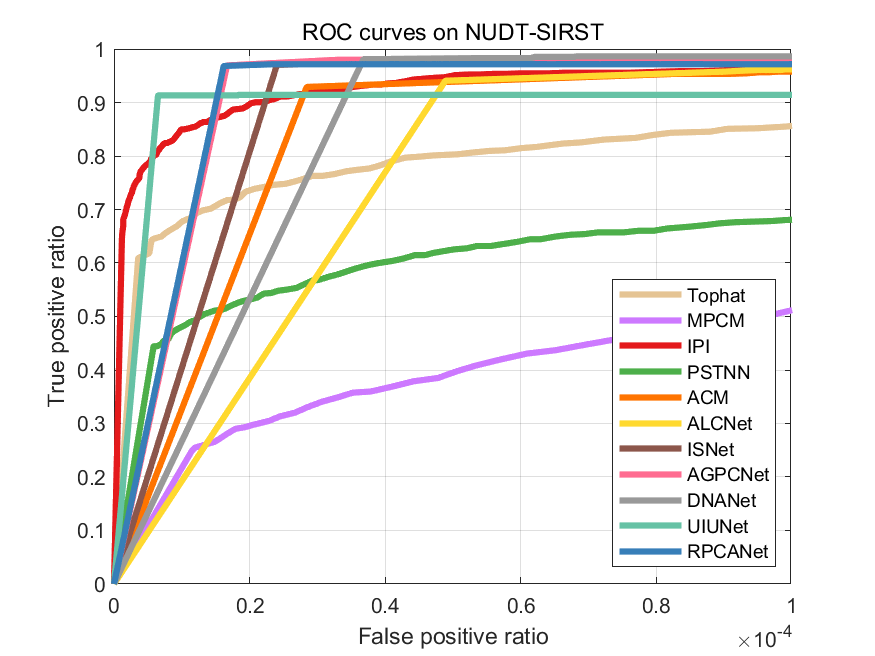}
    \caption{ROC curves of different methods on NUDT-SIRST \cite{li2022dense}.}
    \label{fig:roc}
    \vspace{-0.2cm}
\end{figure}

{\bf Quantitative Results:} To showcase the effectiveness of our network, we compare its detection performance with the SOTA baseline. Table \ref{baseline} demonstrates that RPCANet outperforms most model-driven and data-driven ISTD frameworks in four indicators using fewer parameters. Model-driven methods like IPI excel in target-level metrics but lack pixel-level accuracy. Data-driven networks improve $mIoU$ and $F_1$ scores while preserving the target's shape. However, DL-based algorithms like AGPCNet and UIUNet suffer from overfitting and require more parameters. In contrast, RPCANet combines model-driven priors with accurate object extraction and segmentation guided by data, achieving superior performance with fewer parameters. ROC curves in Fig. \ref{fig:roc} highlight that our model rapidly reaches the upper-left corner and exhibits competitive performance in terms of AUC: RPCANet (0.9857), ACM (0.9830), and UIUNet (0.9477).  Moreover, the last column in Table \ref{baseline} shows that our framework obtains great computational efficiency.

\section{Conclusion}
In this paper, we propose an interpretable framework for infrared small target detection based on deep unfolding networks. We model the ISTD task as a relaxed RPCA problem and solve the optimization steps via network emulations, including the proximal network and sparsity-constrained neural layers. Our scheme produces trustworthy visualizations and outstanding detection results in extensive experiments on various public datasets. The architecture of RPCANet effectively guides neural layers to learn low-rank backgrounds and sparse targets, facilitating the detection tasks in an almost "white box" manner. We hope our findings cloud inspire researchers to explore more interpretable solutions for ISTD problems in the future.
% In this paper, we explore the low rank and sparse representation of infrared small target detection that can be expressed as neural networks and propose a novel interpretable network framework for ISTD. We first model the ISTD task into a relaxed RPCA model and then draw the optimization subproblems. The complex matrix operations in optimization closed-form solutions are solved by network emulations based on the property of Lipschitz continuity, and thus the network is almost "white-box" with decent visualized verification for all modules. Extensive experiments implemented on public datasets prove that the explainable framework maintains a good detection performance and how this architecture guides neural layers to learn backgrounds and finally facilitates the detection tasks. We hope our findings could inspire field scholars to rethink the solution to ISTD problems in a more interpretable way.   
\section*{Acknowledgments}
This work was supported by the Natural Science Foundation of Sichuan Province of China (No.2022NSFSC40574) and National Natural Science Foundation of China (No.61775030, No.61571096).

%%%%%%%%% REFERENCES
{\small
\bibliographystyle{ieee_fullname}
\bibliography{egbib}
}

\end{document}